# Customized Co-Simulation Environment for Autonomous Driving Algorithm Development and Evaluation

Mustafa Ridvan Cantas and Levent Guvenc

Automated Driving Lab, Ohio State University

## Abstract

Deployment of autonomous vehicles requires an extensive evaluation of developed control, perception, and localization algorithms. Therefore, increasing the implemented SAE level of autonomy in road vehicles requires extensive simulations and verifications in a realistic simulation environment before proving ground and public road testing. The level of detail in the simulation environment helps ensure the safety of a real-world implementation and reduces algorithm development cost by allowing developers to complete most of the validation in the simulation environment. Considering sensors like camera, LiDAR, radar, and V2X used in autonomous vehicles, it is essential to create a simulation environment that can provide these sensor simulations as realistically as possible. While sensor simulations are of crucial importance for perception algorithm development, the simulation environment will be incomplete for the simulation of holistic AV operation without being complemented by a realistic vehicle dynamic model and traffic co-simulation. Therefore, this paper investigates existing simulation environments, identifies use case scenarios, and creates a co-simulation environment to satisfy the simulation requirements for autonomous driving function development using the Carla simulator based on the Unreal game engine for the environment, Sumo or Vissim for traffic co-simulation, Carsim or Matlab/Simulink for vehicle dynamics co-simulation and Autoware or the authors' or users' own routines for autonomous driving algorithm co-simulation. As a result of this work, a model-based vehicle dynamics simulation with realistic sensor simulation and traffic simulation is presented. A sensor fusion methodology is implemented in the created simulation environment as a use case scenario. The results of this work will be a valuable resource for researchers who need a comprehensive co-simulation environment to develop connected and autonomous driving algorithms.

## Introduction

Fast development requirements for autonomous vehicles lead algorithm developers to utilize various simulation environments. Simulations are not only faster and cheaper, but also safer as compared to real-world testing. In our application, we are mainly focusing on developing sensor fusion applications that can be utilized in an Advanced Driver Assistant System (ADAS) or Connected and Autonomous (CAV) applications. The initial development of sensor fusion algorithms can be carries out using collected data or in a simulation environment.

One can utilize three primary data resources to develop sensor fusion algorithms. These are the use of: simulation environments, readily available data sets, and custom data sets. Some of the publicly available datasets can be listed as: Kitti Dataset [1][2][3][4], Argoverse Dataset [5], nuScenes [6], Ford AV Dataset [7], Lyft Level 5 Perception Dataset [8], Astyx HiRes2019 [9], Oxford Radar RobotCar [10], and the OSU-ADL Smart Columbus Dataset[11,12]. While these datasets are mostly based on camera and lidar sensors, nuScenes, Astyx HiRes2019, and Oxford Radar RobotCar datasets are the only datasets that have radar data as well. Although the datasets eliminate the time required to collect data, the reliance on data collected by others for a different purpose may limit the algorithm developer in terms of the scenario variety and sensor selection. This can be addressed by creating custom scenarios in a simulation environment or collecting a custom set of data. Some of the available photorealistic simulation environments for autonomous vehicles can be listed as: Carla [13], LGSVL [14], Airsim [15]. One of the advantages of the simulators is having the ground truth for the target vehicle. Considering the scenario customization flexibility, repeatability, and accurate ground truth information advantages of simulations, a customized simulation environment is constructed for our research purpose. While our main target is to use the developed environment for sensor fusion, other functionalities, and potential extensions of the simulation environment are also presented.

The following sections are organized as follows. Firstly, the simulation environment and its potential extension capabilities are explained. Secondly, the LiDAR 3D object detection method and its implementation are demonstrated. Then, a brief introduction to V2X is presented. Afterwards, a sensor fusion method that is used with raw data generated by the simulator is outlined. As a more interesting use case scenario, a Collaborative Perception application is introduced. Then, finally, a summary of the paper with potential improvements and future work is presented.

## Simulation Environment

As mentioned in the introduction part, there are many commercial and non-commercial simulators that can be used to develop autonomous vehicles. In this work, Carla is selected because of its flexible architecture, which allows extending the simulator capabilities. In Figure 1, one can see how this simulator can be extended with other simulation environments. Most of the shown co-simulation capabilities are supported by Carla. For example, ROS bridge, Autoware, Sumo, and Vissim co-simulation environments are directly supported by Carla. Also, one can create new maps and simulation environments with Matlab RoadRunner and import them to Carla. Considering that Carsim, Matlab, and Carla software can communicate with Unreal Engine, it is possible to create co-simulation environments where the vehicle dynamics are simulated in Matlab or Carsim in a Carla simulation. Finally, by using the property of MATLAB for running Python scripts, one can utilize MATLAB - Carla co-simulation. For the presented work, we decided to use a subset of the shown simulation architecture. Since the demonstrated work will mostly focus on sensor fusion applications, we are primarily interested in sensor data. Therefore, our co-simulation subset consists of Matlab – Carla simulation environment with Sumo co-simulation. While the Matlab interface is used to transfer sensor data to Matlab, with the addition of Sumo co-simulation we can also inject traffic into our simulation environment.

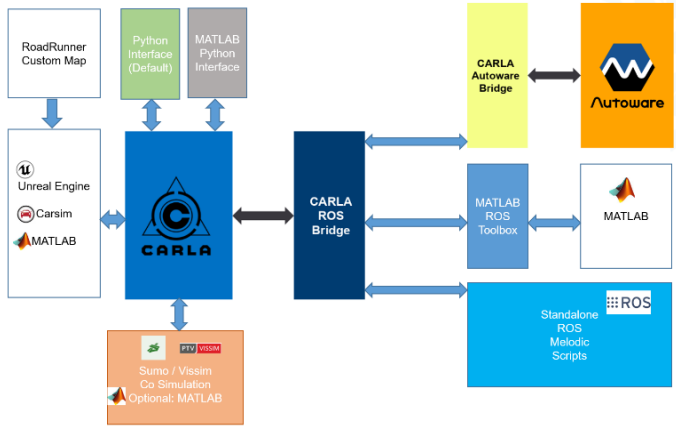

Figure 1. Carla Simulator extension diagram

*Python Interface*

The main interface for the Carla simulator is Python scripts. By using PythonAPI, it is possible to select maps, change the weather, spawn vehicles, pedestrians, and other actors, control the actors, and more. In our application, we set our simulation environment using our custom Python script based on the target task, such as sensor fusion or car following.

*ROS Interface*

The Robot Operating System (ROS) is a robotics middleware which eases managing multiple processes in a system. It provides services to exchange common messages between processes and lets developers control hardware. ROS also allows users to create packages that can be reusable in different systems. The latest version of Carla (9.10) supports both ROS and ROS2 by having a ROS bridge. In this manuscript, the ROS interface is used to connect to Autoware and MATLAB to collect data and interact with the Carla simulator. Autoware can be utilized for path planning, localization, object detection, and lateral & longitudinal control tasks. The implemented Carla-Autoware co-simulation is presented in Figure 2.

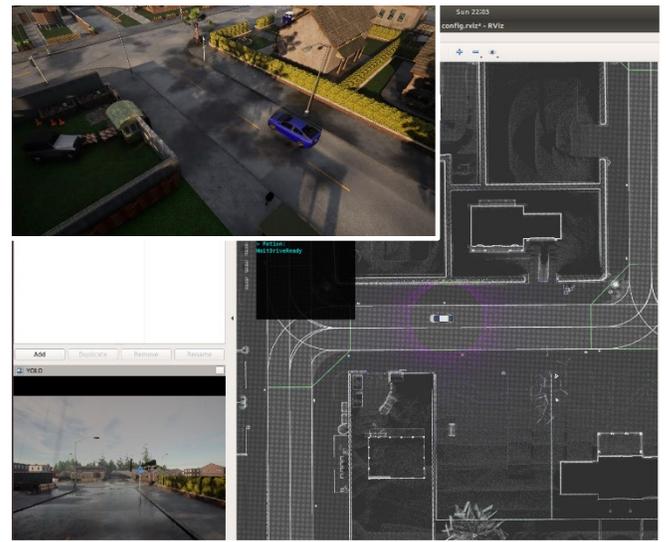

Figure 2. Top Left: Carla simulator interface. Background: Snapshot of Rviz ROS interface to Carla Autoware integration showing point cloud map, HD map, LiDAR, and camera sensor outputs.

*MATLAB Interface*

As explained in previous sections, there are two main interfaces for Carla, which are the ROS and Python script interfaces. Since Matlab supports both of these interfaces, Matlab-Carla simulation can be realized by using one of them. While Python scripting is supported directly within MATLAB scripts, to interact with the Carla ROS interface, the MATLAB ROS toolbox is used. We built our MATLAB Carla integration based on [16]. One should note that the MATLAB ROS integration is limited by the implemented features in the Carla ROS Bridge. On the other hand, the Python interface in MATLAB provides much more flexibility for customizing the simulation environment.

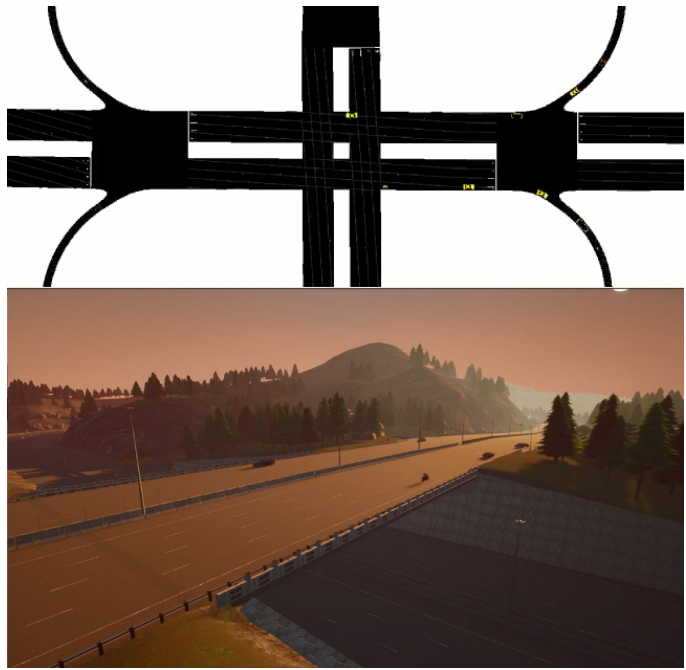

Figure 3. Carla Sumo Integration

*Traffic Simulation*

Carla creates traffic around the host vehicle by randomly spawning vehicles. For evaluating the potential mobility or fuel-efficiency benefits of connected and autonomous vehicles, enhanced traffic simulations can be used to create realistic traffic simulation. Some of the examples of these applications can be seen in [17-19]. Among many of the micro traffic simulators, Sumo and PTV Vissim are the most commonly preferred simulators being used. They are also supported by Carla. In Figure 3, a screenshot from our sample Sumo-Carla co-simulation implementation can be seen. The Sumo road network is shown on top and the corresponding Carla road environment is shown at the bottom in Figure 3.

*Vehicle Dynamics*

Vehicle dynamics in Carla is modeled using the NVIDIA PhysX Vehicle, which does not accurately represent the vehicle dynamics. However, to develop and analyze control algorithms for AVs, the realistic vehicle dynamics model is required to develop algorithms which can handle the dynamic behavior of vehicle under consideration [20-26]. In some cases, having the full vehicle model with powertrain also helps to develop more accurate fuel-efficiency algorithms [27-30]. To address this, it is proposed to extend the simulation by integrating it with CarSim, as shown in [31] or MATLAB Vehicle models. Models developed in MATLAB can be generated using its recent Vehicle Dynamic Toolbox, or it can be a completely custom model. In the Vehicle Dynamics co-simulation environment, vehicle dynamics would be simulated in Carsim or Matlab, while the photo-realistic simulation environment and other simulated sensor data come from Carla.

## LiDAR 3D Object Detection

In the sensor fusion part, LiDAR detections will be fused with V2X messages, which requires tracks from each sensor modality. Therefore, this part of the paper will explain how object detection and tracking is achieved. The connection between Carla and Matlab is established using Python scripting in Matlab. Then each LiDAR frame is processed to detect and track objects. While in the literature, there are many deep learning-based applications that exist for LiDAR 3D object detection, in this work, a simpler method presented in [32] is used. In this method, the bounding box detection is achieved by processing the LiDAR frame in three steps:

1- Points outside the region of interest are removed.
2- Ground plane points are removed.
3- Points representing obstacles are clustered and a bounding box is created.

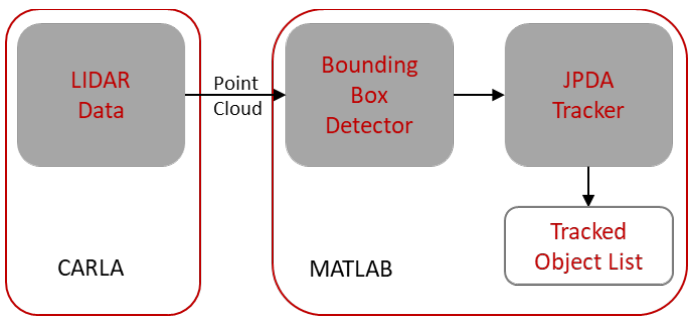

Figure 4. Carla Matlab co-simulation LiDAR 3D object detection

After the bounding box is detected a Joint Probability Data Association (JPDA) Multi Object Tracker from the Matleb Sensor Fusion Toolbox is used to detect and track targets. The Carla-Matlab co-simulation environment structure for LiDAR 3D Object Detection is shown in Figure 4. Detection results for the 3D object detection is shown in Figure 5.

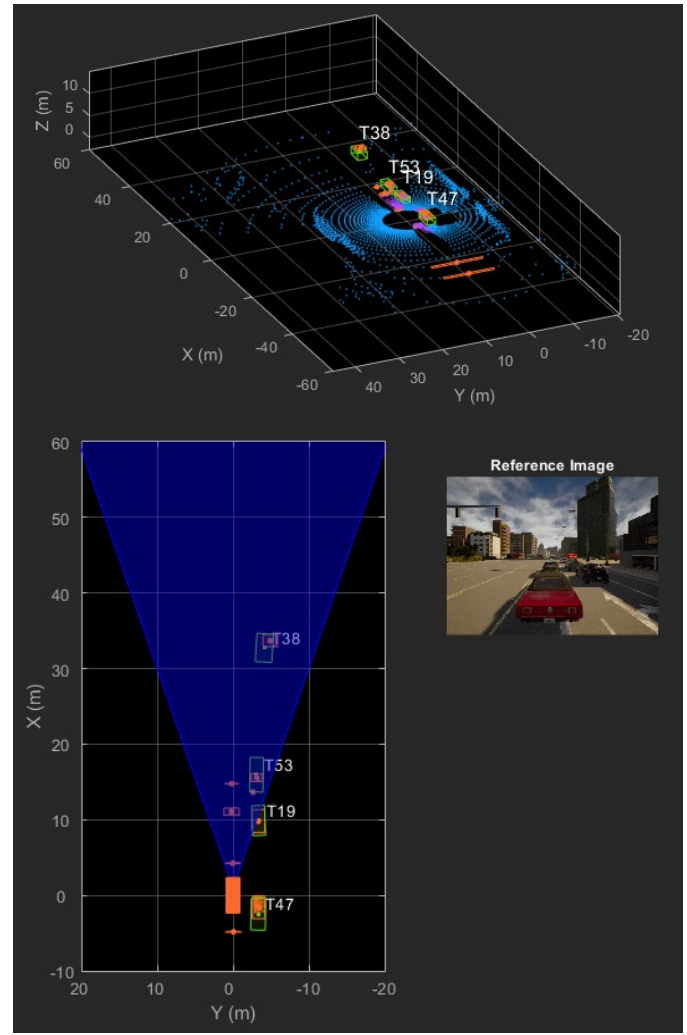

Figure 5. LiDAR 3D object detection and tracking

## V2X Coomunication

The use of Vehicle-to-Everything (V2X) communication as an extra sensor increases the total field of view of the vehicle significantly. Extension of the field of view helps to eliminate accidents due to the occlusion of obstacles/objects. Also, sharing the detected objects through V2X communication can increase the effectiveness of V2X applications [33-36]. The effect of V2X communication is especially significant when some of the vehicles do not have V2X units. By broadcasting the range sensor detection through V2X communication, the interaction of the deployed vehicles with other vehicles would be increased significantly [34]. In [34], the authors tested the collaborative perception with a vehicle, which has a camera and Dedicated Short Range Communication (DSRC) onboard unit (OBU). They showed that using the collaborative perception, V2V applications, such as Left Turn Assist (LTA), Intersection Movement

Assist (IMA), and Blind Spot Warnings(BSW) still perform well even with the delay caused by the perception information processing.

While many of the applications focus on sensor fusion for Camera, LiDAR, and Radar, in this paper, we are proposing to fuse Vehicle-to-Vehicle (V2V) communication tracks with LiDAR tracks. The SAEJ2735 [37] standard proposes Basic Safety Messages (BSM) to be broadcast at a 10 Hz rate from all connected vehicles. BSM consists of two parts. In our application, we are mainly utilizing a subset of the Part I messages. While the Carla simulator does not have integrated BSM messages, in our simulation environment, we created a custom connected vehicle that creates the required subset of BSM for object tracking. Implemented messages in Part I can be listed as: time, latitude, longitude, altitude, speed, acceleration, and heading. While in LiDAR detections, it was required to track the vehicles with an algorithm, for V2V communication, we can exploit published ID information to associate broadcast BSMs.

## Sensor Fusion

For an autonomous vehicle, it is crucial to have an accurate, robust, and real-time scene perception architecture. Although great advances were achieved in automotive scene perception, the existing methods do not offer an optimal solution for fusing data received from different sensor modalities. Therefore, the automotive industry and researchers are still in search of better sensor fusion architectures. Also, in the literature, reported work on sensor fusion of new automotive sensors such as V2X is limited. In this manuscript, we are presenting a customized simulation environment to be used for the development of connected vehicle sensor fusion applications. While many sensor fusion applications can be initially developed using readily available datasets, for the fusion of V2X with other sensors, there is no publicly available dataset. Thus, for our sensor fusion development task, the developed simulation environment will be employed. In the created simulation environment, each connected vehicle is equipped with an On-Board Unit (OBU), which broadcasts BSM. The host vehicle is equipped with LiDAR, radar, camera, and OBU. Sensor data from the host and target vehicles are received using a Python script in Matlab. Then, the received data is fed to the sensor fusion module as shown in Figure 6. To demonstrate the effectiveness of the simulation environment, LiDAR detections will be fused with BSMs. Similar to the previous section, LiDAR detections are obtained by creating bounding boxes for obstacles detected in Point Cloud Data (PCD). Then, the list of detected objects and received BSM messages are passed to the JPDA Tracker. The JPDA tracker then assigns each detected object from each sensor to a track. This can be an existing track or a new track. Finally, fused tracks will be the output of the sensor fusion module.

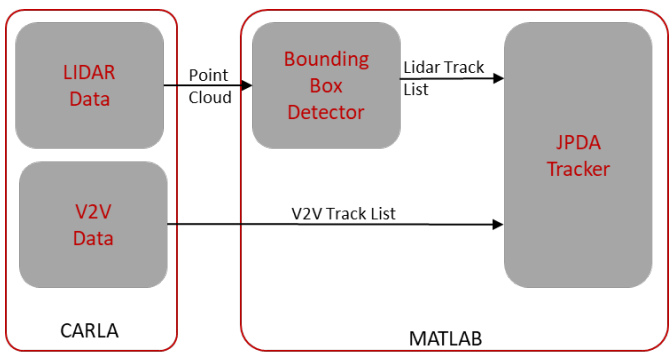

Figure 6. Sensor Fusion Simulation Architecture

## Collaborative Perception

The perception of the surrounding vehicles and environment information using only range sensors is challenging for vehicles. It is impossible to detect vehicles outside of the Field-of-View (FOV) of the range sensors or vehicles which are far away. This research aims to maximize the awareness of all the vehicles in a mixed traffic environment with the use of range sensors and communication technologies. The mixed traffic consists of vehicles without any detection or communication capability (No Sensing), Connected Vehicles, and Connected Vehicles with range sensors, such as Connected and Autonomous Vehicles. In Figure 7, the benefit of the Collaborative Perception approach is presented. Vehicles 4, 13, 14 and 15 are connected vehicles without range sensors and broadcast only their own BSM values. The host vehicle HV and vehicles 3, 9, 19 are connected vehicles that are also equipped with range sensors. They share not only their own BSM but also generate and share BSM data for the detected vehicles which do not have a V2X unit. Therefore, now the HV that can perceive vehicles 1 and 2 with its range sensors also has the BSM information of vehicles 5, 6, 10, 11 ,12, and 16 which is generated by the other communicating vehicles with range sensors (3, 9, 19). Vehicles 7, 8, 17 and 18 are not perceivable even with collaborative perception but they are not relevant in location as far as the HV is concerned. As can be seen from this simple illustration, the situational awareness of the HV is improved significantly, benefiting from collaborative perception.

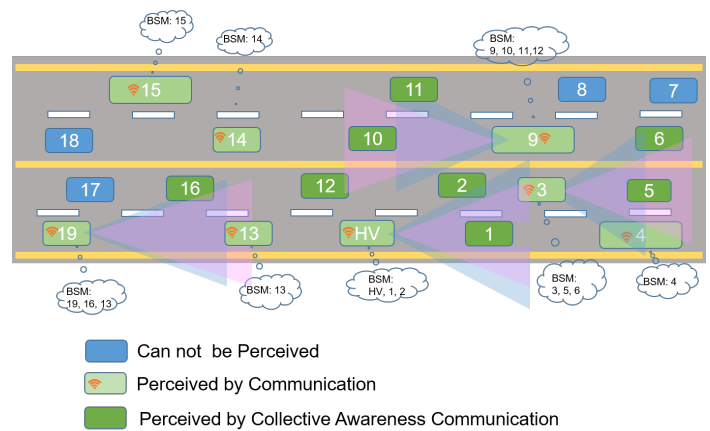

Figure 7. Collaborative Perception in a mixed traffic simulation environment

Motivated by this, we demonstrate a collaborative perception scenario as a use case for the customized Carla Simulation environment, as shown in Figure 8. In this scenario, a pedestrian is crossing the road but it is not visible to the host vehicle. The pedestrian is obstructed by the truck, which resides between the host vehicle and the pedestrian. The truck equipped with a collaborative perception module can increase the situational awareness of the host vehicle and help it to avoid a serious accident. In Figure 9, one can see the coordinate transformation process of perceived targets in the vehicle coordinate system to the global coordinate system and BSM creation for detected objects. The fused detection list will be the sensor fusion output of range sensors like radar, camera, and LiDAR.

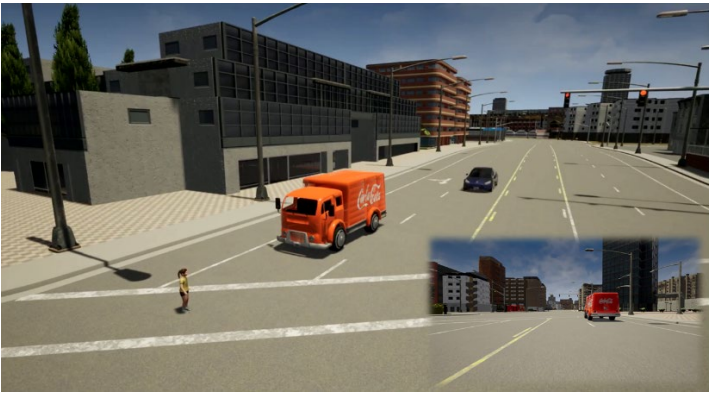

Figure 8. Right bottom foreground: Host vehicle camera image, Background: scenario image from an external camera showing view that is obstructed by truck.

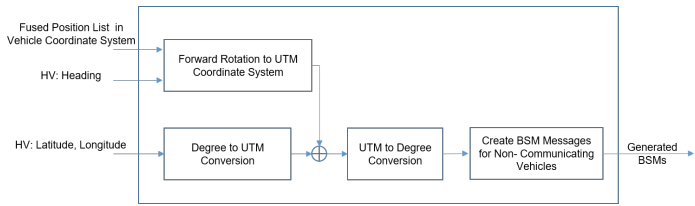

Figure 9. Coordinate transformation & BSM generation process.

In the presented application, situational awareness is increased by fusing range sensor measurements with received BSM messages. Here the received BSM messages include self-reporting BSMs and BSMs shared by the collaborative perception unit of another vehicle.

## Summary/Conclusions

It is crucial for automakers to implement safe, reliable driving assistant technologies and autonomous driving functions. Since it is expensive and time-consuming to develop algorithms in a real-world environment, HIL, MIL, or SIL simulations are heavily employed in the industry. Similarly, in this work, a customized simulation environment for developing sensor fusion applications and their applications is presented. The customized environment enables us to simulate V2X applications, such as collaborative perception. This simulation environment can be further improved by integrating vehicle dynamic models from CarSim or the Matlab Vehicle Dynamic Toolbox. The co-simulation environment and approach presented in this paper can be used to aid the development and evaluation of different control, automotive control, ADAS and connected and autonomous driving methods [38-76].

## Definitions/Abbreviations

| | |
|---|---|
| **V2V** | Vehicle to Vehicle |

| | | | |
|---|---|---|---|
| **AV** | Autonomous Vehicle | **OBU** | On Board Unit |
| **ADAS** | Advanced Driver Assistance System | **JPDA** | Joint Probability Data Association |
| **CAV** | Connected and Autonomous Vehicle | **PCD** | Point Cloud Data |
| **V2X** | Vehicle to Everything | **FOV** | Field of View |
| **BSM** | Basic Safety Message | | |
| **DSRC** | Dedicated Short Range Communication | | |